\documentclass{article}

\usepackage[left=3cm, right=3cm, top=3cm, bottom=3cm]{geometry}

\usepackage{graphicx}

\usepackage{here}

\usepackage{amsmath,amssymb}
\usepackage{amsfonts}

\title{Data Cleansing for Deep Neural Networks \\ with Storage-efficient Approximation of Influence Functions}

\author{Kenji Suzuki\thanks{Kenji.B.Suzuki@sony.com}, Yoshiyuki Kobayashi, Takuya Narihira \\
 {\large Sony Group Corporation, Japan}
}

\date{}

\begin{document}

\maketitle

\begin{abstract}
    Identifying the influence of training data for data cleansing can improve the accuracy of deep learning. An approach with stochastic gradient descent (SGD) called SGD-influence to calculate the influence scores was proposed, but, the calculation costs are expensive. It is necessary to temporally store the parameters of the model during training phase for inference phase to calculate influence sores. In close connection with the previous method, we propose a method to reduce cache files to store the parameters in training phase for calculating inference score. We only adopt the final parameters in last epoch for influence functions calculation. In our experiments on classification, the cache size of training using MNIST dataset with our approach is 1.236 MB. On the other hand, the previous method used cache size of 1.932 GB in last epoch. It means that cache size has been reduced to 1/1,563. We also observed the accuracy improvement by data cleansing with removal of negatively influential data using our approach as well as the previous method. Moreover, our {\it simple} and {\it general} proposed method to calculate influence scores is available on our auto ML tool without programing, Neural Network Console. The source code is also available.
\end{abstract}

\section{Introduction}
There is a growing interest in the fairness, accountability, and transparency of machine learning with the outbreak of AI ethical issues. The machine learning is called black box systems because of very complex networks with large amount of neurons. Understanding the behavior of neural networks remains a significant challenge. Explainable AI can be used not only for AI ethics applications, but also for deep neural network debugging applications, and has the potential to extract the potential of deep neural networks. Developers can analyze interpretations to debug models. Researchers have developed some methods to open the black box, such as LIME~\cite{LIME} and Grad-CAM~\cite{GradCAM} to visualize the basis of judgement in classification. These methods help humans to understand the reason of judgement. 

\subsection{Motivation}
Although these methods can indicate the basis of judgement with heat map or super pixels, we would like to improve accuracy without adding new data as next step. It is very crucial to suggest what to do in explainable AI. Recently some methods for data cleansing calculating influence of data to deep neural networks have been proposed~\cite{InfluenceFunctions}~\cite{SGD}. 

The influence scores can specify negative influential score to deep neural networks. These influence methods help developers to clean a large amount of data for high-quality data without domain knowledge for labeling. Especially, SGD-influence~\cite{SGD} using stochastic gradient descent (SGD) proposed by Hara {\it et al.} does not require the loss function to be convex and an optimal model to be obtained. However, the SGD-influence needs a large mount of cache files to store parameters in training phase. We are motivated to calculate influence with storage-efficient approximation, and to implement it on Neural Network Console~\footnote{https://dl.sony.com/} that is our auto ML tool. Therefore, we study how to reduce the size of cache with equivalent accuracy improvement.

\section{Our Contributions}
 We propose a method of SGD-influence with cache reduction. It enables us to implement the algorithm on Neural Network Console, our auto ML tool, that we developed. In this paper, we propose a novel cost-effective data influence algorithm for data cleansing in deep learning. Our key contributions are summarized as follow:
\begin{enumerate}
    \item We propose cache reduction algorithm for SGD-influence. The algorithm is based on SGD-influence~\cite{SGD} proposed by Hara {\it et al.} The original work needs a large amount of HDD cache between training phase and inference phase. We adopt only the last parameters in last epoch for influence functions calculation.
    \item We demonstrate that our proposed method can effectively improve accuracy by removing the negative influential data.   
    \item As a result, it enables us to calculate influence scores without large HDD cache. Therefore, we have implemented this proposed algorithm on our auto ML tool called Neural Network Console which is available. Using this software, the influence scores are easily obtained without programming.
\end{enumerate}

\section{Related Work}
In this section, we provide an overview of existing influence methods. 
Classic technique from statistics~\cite{Cook} has shown estimated influence analytically calculated. The influence functions for machine learning models have not seen widespread use in spite of study in statistics. Koh and Liang~\cite{InfluenceFunctions} provided tractable approximations of influence functions which characterize the influence of each data in terms of change loss. However, the method requires the loss function to be convex and an optimal model, which cannot always be satisfied in deep learning. Hara {\it et al.}~\cite{SGD} provided the influential instances by retracing the steps of the SGD without these limitations.

\subsection{Baseline Method}
Our approach is based on SGD-influence~\cite{SGD} for deep neural networks prediction as a baseline. The SGD-influence is based on counterfactual SGD where one instance is absent. The $t$-step of the counterfactual SGD with $j$-th instance $z_{j}$ absence is denoted as $ \theta_{-j}^{[t+1]} \leftarrow \theta_{-j}^{[t]} - \frac{\eta_{t}}{|S_{t}|} \sum_{i\in S_{t}\backslash\{j\}}g(z_{i};\theta^{[t]}_{-j}) $ 
where $S_{t}$ denotes the set of instance indices used in the $t$-th step, $\eta_{t}$ is learning rate, $\theta$ is the parameters of the model, $z$ is an instance. Let $ g(z;\theta):= \nabla_{\theta}l(z;\theta)$, where $l$ is loss function.
SGD-influence refers to the parameter difference $\theta^{[t]}_{-j} - \theta^{[t]}$ as the instance $z_{i} \in D$ at step $t$. Linear influence estimation (LIE) is estimated liner influence $L^{[T]}_{-j}(u^{[t]}):=\langle u^{[t]},\theta^{[T]}_{-j}-\theta^{[T]} \rangle$ for a given query vector $u^{[T-1]} \in \mathbb{R}^{p}$ and the SGD-influence after $T$ SGD steps.

The SGD-influence consists of training phase and inference phase. The tuple of instance indices $|S_{t}|$, learning rate $\eta_{t}$, and parameters ${\theta}^{[t]}$ are stored in training phase. The stored information is retraced and $u^{[t]}$ in each step is computed. $H^{[t]}$ is the Hessian of the loss on the mini-batch $S_{t}$.   

They evaluated both trace back all training epochs and approximate version of the trace back only last one epoch which is storage friendly, which counterfactual SGD is used in last epoch. They observed that the approximate version is also effective for data cleansing. As a baseline method, we study comparison of the approximate version with our approach. 

\begin{figure}[t]
    \centering
    \includegraphics[width=1.0\linewidth]{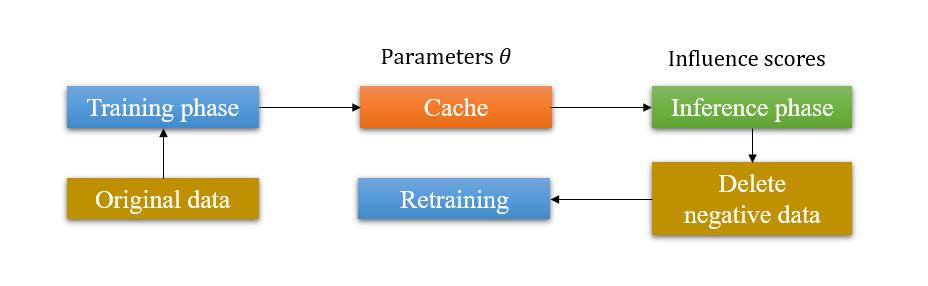}
    \caption{Workflow of data cleansing. To compute influence scores, there are training phase, cache, and inference phase. The cache stores parameters from the training phase. To improve accuracy, there are stages of deleting negative data and retaining. }
    \label{Fig.1}
\end{figure}

\section{Proposed Method}
Figure~\ref{Fig.1} shows workflow of data cleansing. To compute influence scores, there are training phase, cache, and inference phase. To improve accuracy, there are stages of deleting negative data and retaining. 

In order to reduce the amount of the parameters from the training phase, we propose to store only final parameters in last epoch from training phase. It means that cache size has been reduced to $1/T$, where $T$ is SGD steps. The previous study of SGD-influence as a baseline stores each SGD step $T$ in training phase. Here we assume that the final parameters make important rules for models. Therefore, we store the final parameters in last epoch and load the final parameters to calculate influence as shown in Algorithm 1 and 2. 

\begin{figure}[h]
    \centering
    \includegraphics[width=1.0\linewidth]{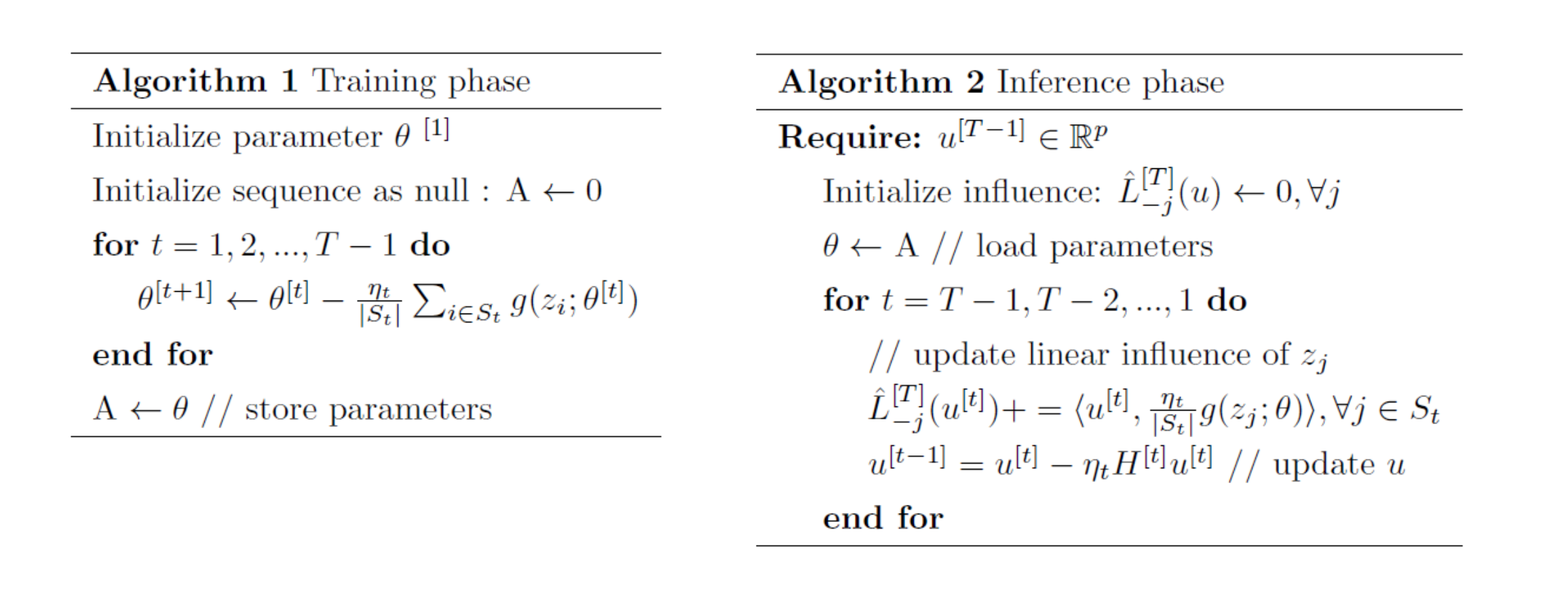}
\end{figure}

This may not be exact computation, but we regard the method as an approximation because there are little difference among parameters of each SGD step in last epoch. For example, when the number of dataset is 50,000 and the batch size $|S_{t}|$ is 32, there are $T=1,563$ SGD steps in the last epoch. We use only final step instead of the all parameters in each step. Therefore, we can reduce the cache size to 1/1563 ($1/T$). Since the number of data $N$ is $T \times |S_{t}|$, cache reduction is $1/T \risingdotseq |S_{t}|/N$. This relationship indicates that it is useful for calculating large amounts of data $N$ in large-scale calculations. 

This proposal is very {\it simple} and {\it general}. In practical use of deep learning, computation cost is an important factor. This proposed approximation is very useful for developers.

\section{Evaluations}
We employ Neural Network Libraries~\cite{NNL2}which is our developed deep learning framework under CUDA Tool kit 10.2 and cuDNN 8.0. Neural Network Libraries are available as an OSS~\footnote{https://nnabla.org}. The source code using the Neural Network Libraries in our experiment is available at Sony AI research code repository~\footnote{https://github.com/sony/ai-research-code}.   

In order to evaluate the cache reduction and data cleansing performance, we conducted image classification. We used two datasets, MNIST~\cite{MNIST} and CIFAR10~\cite{CIFAR10}, to evaluate that models of deep neural networks are effectively improved by removing negative influence data. The deep neural networks consist of 6 conventional convolution layers.  

\subsection{Cache Reduction}
The table 1 shows cache size of the all parameters, the previous method of Hara {\it et al.}, and our proposed method for dataset of MNIST and CIFAR10. We use 50,000 and 40,000 training data for MNIST and CIFAR10, respectively. All parameters are accumulated through epoch $k=$ 20. In the previous method~\cite{SGD} that use the parameters in the last epoch, it is necessary to use 1.932 GB and 1.545 GB in HDD cache for MNIST and CIFAR10, respectively. Our method requires cache of 1.236 MB for MNIST and CIFAR10. We set batch size $|S_{t}|=$ 32, so that MNIST and  CIFAR10 have $T=1,563$ steps and $T=1,250$ steps, respectively. The cache reduction is $1/T$, where $T$ is the number of SGD steps. This result indicates the cache size reductions are 1/1,563 and 1/1,250 for MNIST and CIFAR10, respectively. 

\vspace{5mm}
\begin{center}
Table 1 : Cache size of the parameters in training 
\begin{tabular}{cccc}
    \hline
        Methods         &      Cache        &    \multicolumn{2}{c}{Cache size  (GB) }   \\
                             &                        &    MNIST & CIFAR10    \\
    \hline \hline
    All parameters &  $\theta \times T \times k$  & 38.64  & 30.90\\
    Hara {\it et al.}&  $ \theta \times T $ & 1.932 & 1.545 \\
    Ours & $ \theta$ & 0.001236  & 0.001236 \\
    \hline
\end{tabular}
\end{center}
\vspace{5mm}

\subsection{Data Cleansing Performance}

\begin{figure}[ht]
    \centering
    \includegraphics[width=0.8\linewidth]{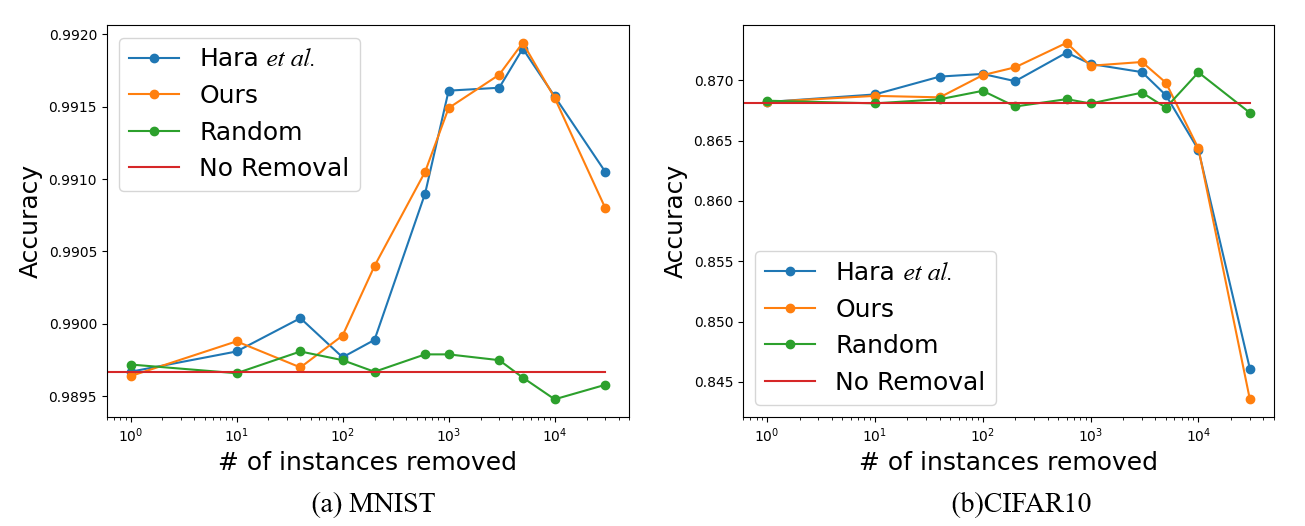}
    \caption{The accuracies on the test dataset after data cleansing deleted data of negative influence scores for (a) MNIST and (b) CIFAR10. Comparison of Hara {\it et al.}, our proposed method, randomly removed, and no removal.}
    \label{Fig.2}
\end{figure}

In order to check the performance of data cleansing, we adopted the original SGD-influence (denoted Hara {\it et al.}) and random data removal as baselines. We set leaning rate $lr=$ 0.05, the the number of epoch $k=$ 20, and batch size $|S_{t}|=$ 32. We randomly selected 10,000 data for validation data and used the remaining data as training dataset from the original dataset. We repeated the experiments 10 times changing different random seeds and the split between training and validation dataset. Another 10,000 data are set for testing accuracy.

Figure~\ref{Fig.2} shows the accuracy on the test dataset after data cleansing for (a) MNIST and (b) CIFAR10. The data cleansing involves first determining the influence of each data and then removing the data that would negatively affect it. In other words, Figure~\ref{Fig.2} shows how the accuracy changes as we remove the top-{\it n} negative data. Increasing accuracy accompanied by removing top-{\it n} negative data is an evidence of data cleansing. We observed our method has the significant performance of data cleansing for both MNIST and CIFAR10 datasets as well as Hara {\it et al.}. On the other hand, we do not observe the significant accuracy improvement by randomly removed data. In the MNIST experiment with batch size $|S_{t}|=$ 32, the accuracy peaked at 5,000 data removals. Since the the number of original training data is 50,000, it shows that the peak of accuracy is 10\% of the data, 5,000, removed from the training data, 50,000. Then the standard deviation $ \sigma $ is 0.000837. Since we repeated the experiments 10 times changing different random seeds, the accuracies in Fig.~\ref{Fig.2} are the mean value. The batch size dependence experiments from $|S_{t}|=$ 32 to 256 show data cleansing behavior at any batch sizes~\footnote{See Appendix for the full results.} .    

\section{Applications}
We have implemented this proposed algorithm on Neural Network Console which is available for users. 
\begin{figure}[b]
    \centering
    \includegraphics[width=0.75\linewidth]{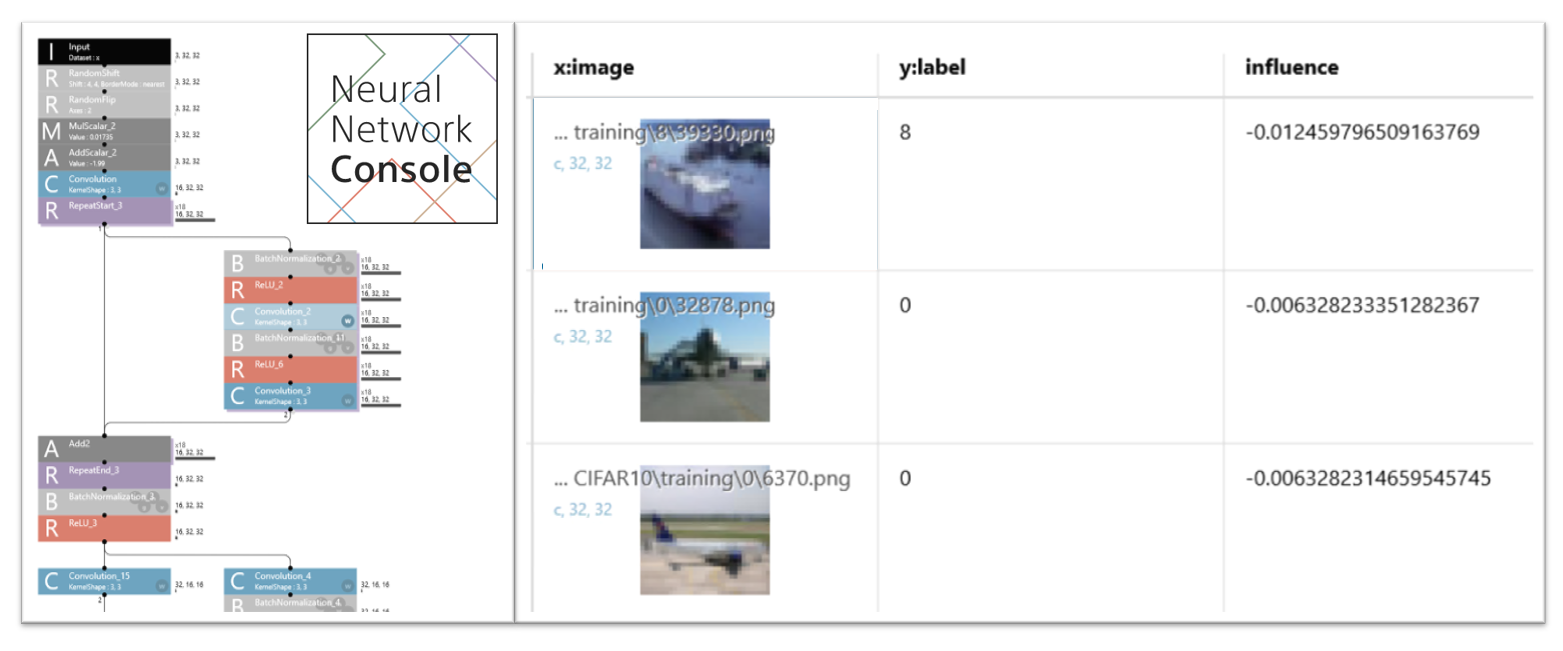}
    \caption{The left part shows screen shot of a part of ResNet-110 on Neural Network Console. The right part shows calculated influence score for each training data of CIFAR10 with ResNet-110. Our proposed method is available on our auto ML tool, Neural Network Console.}
    \label{Fig.3}
\end{figure}
The software runs on Windows 10 or cloud service with GPU, {\it e.g.} NVIDIA V100. It is our auto ML tool by GUI, which users can easily use. Users can compute deep learning without programming. Here we provide some plugins of explainable AI on Neural Network Console. The left part in Fig.~\ref{Fig.3} shows screen shot of a part of ResNet-110 on Neural Network Console. This graphical networks help user visually understand. The screen shot at right part in Fig.~\ref{Fig.3} shows calculated influence score for each training data of CIFAR10 with ResNet-110. In addition, this Neural Network Console has a variety of explainable AI algorithms as plugins as well as this SGD-influence, which are LIME~\cite{LIME} and Grad-CAM~\cite{GradCAM} to visualize the judgement basis in image classification as well. User can easily examine these out-of-state explainable AI with GUI on Windows 10 or cloud service.

\section{Conclusion}
Our {\it simple} and {\it general} method presented here, cache reduction of SGD-influence, is a step toward cleansing data in deep neural networks. The significant cache reduction helps developers conduct data cleansing with less computation storage. This result enables us to implement of influence functions calculation by SGD-influence on our auto ML tool, our Neural Network Console. We proved that our method is an effective approximate way to improve accuracy by  removing negative data as well as the previously proposed SGD-influence.

\subsection*{Acknowledgements}
We would like to thank Associate Professor Satoshi Hara of Osaka University for fruitful advice and discussion. We would like to thank Masato Ishii of Sony Corporation of Japan for helpful discussion, and to thank Andrew Shin of Sony Corporation of Japan for reviewing this paper, and to thank Yukio Oobuchi of Sony Corporation of Japan for technical support.

\bibliographystyle{unsrt}
\bibliography{DataCleansing_arXiv_V2}

\newpage

\renewcommand{\thesection}{\Alph{section}}
\setcounter{section}{0}

\section{Full Results}
In this section, we provide additional experimental results to understand the effect of data cleansing. 

\subsection{Standard Deviation}
Since we repeated the experiments 10 times changing different random seeds, the accuracies with standard deviation $ \sigma $ in Fig.~\ref{Fig.4}. In the both MNIST and CIFAR10 experiments with batch size $|S_{t}|=$ 32, we observed that the accuracies have increased with error bars, which mean average $\pm$ standard deviation $ \sigma $. Compared with previous method of Hara {\it et al.}, we note that our proposed method performed similar behavior. The observation suggests that our proposed method with storage-efficient approximation of influence functions should be useful. 

\setcounter{figure}{3}

\begin{figure}[h]
    \centering
    \includegraphics[width=0.9\linewidth]{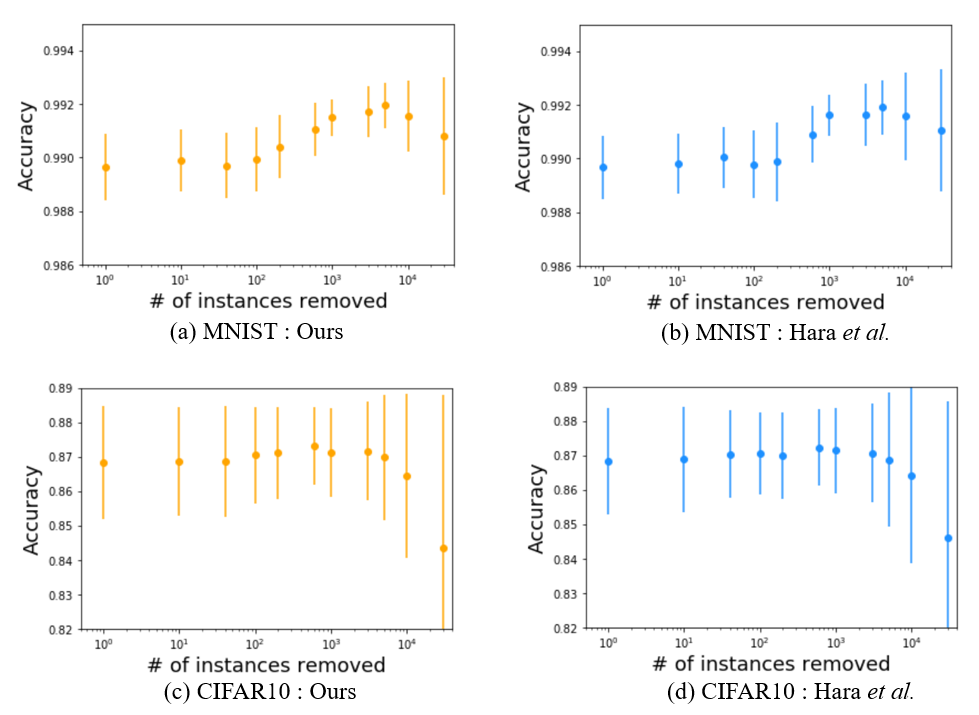}
    \caption{The accuracies on the test dataset after data cleansing deleted data of negative influence scores for MNIST and CIFAR10. Comparison of our proposed method and Hara {\it et al.}. The average accuracies on the test set after data cleansing with 10 experiments. The error bars mean average $\pm$ standard deviation $ \sigma $.}
    \label{Fig.4}
\end{figure}

\subsection{Batch Size Dependence}
\begin{figure}[ht]
    \centering
    \includegraphics[width=0.9 \linewidth]{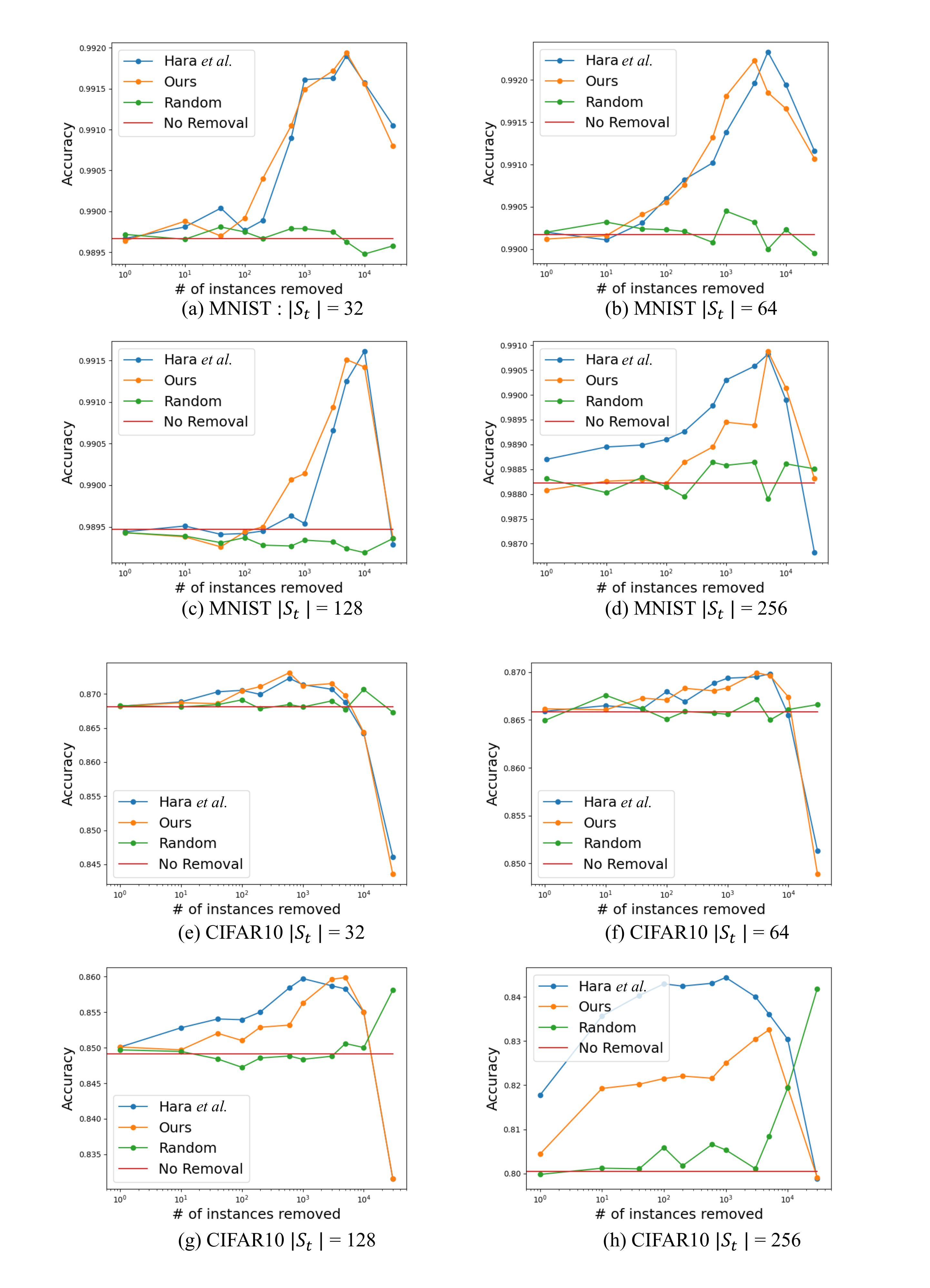}
    \caption{Batch size dependence of the accuracies on the test dataset after data cleansing that deletes data of negative influence scores for MNIST from (a) to (d) and CIFAR10 from (e) to (h). Comparison of Hara {\it et al.}, our proposed method, randomly removed, and no removal.}
    \label{Fig.5}
\end{figure}

Figure~\ref{Fig.5} shows the full results of batch size dependence of the accuracies on the test dataset after data cleansing that deletes data of negative influence scores. It is clear evident that accuracies have increased after data cleansing with our proposed method for various batch sizes in MNIST and CIFAR10 experiments. These results of batch size dependence experiments from $|S_{t}|=$ 32 to 256 confirm that our proposed method can effectively suggest inference functions calculation for data cleansing.

\end{document}